# Robust Degraded Face Recognition Using Enhanced Local Frequency Descriptor and Multi-scale Competition

Guangling Sun, Guoqing Li, Xinpeng Zhang

School of Communication and Information Engineering, Shanghai University, Shanghai, China

sunguangling@shu.edu.cn

*Abstract*—Recognizing degraded faces from low resolution and blurred images are common yet challenging task. Local Frequency Descriptor (LFD) has been proved to be effective for this task yet it is extracted from a spatial neighborhood of a pixel of a frequency plane independently regardless of correlations between frequencies. In addition, it uses a fixed window size named single scale of short-term Frequency transform (STFT). To explore the frequency correlations and preserve low resolution and blur insensitive simultaneously, we propose Enhanced LFD in which information in space and frequency is jointly utilized so as to be more descriptive and discriminative than LFD. The multi-scale competition strategy that extracts multiple descriptors corresponding to multiple window sizes of STFT and take one corresponding to maximum confidence as the final recognition result. The experiments conducted on Yale and FERET databases demonstrate that promising results have been achieved by the proposed Enhanced LFD and multi-scale competition strategy.

*Keywords*- *face recognition; low resolution and blurred; Enhanced LFD; multi-scale competition; frequency correlation;*

## Ⅰ. INTRODUCTION

Due to a wide range of potential applications as well as academic challenges, face recognition has attracted much attention during the last decade. Despite great progress has been made in design of scheme robust to expressions and aging of subjects, partial occlusions, illuminations and inaccurate registrations, most of them aimed at recognizing faces in high quality image. Once coping with degraded images caused by such as blur, low resolution, noise etc, the performance will decline dramatically. Hence, in this paper, we will focus on robust blurred and low resolution face recognition.

There roughly exist three categories of frameworks in literature to handle face recognition from blurred and low resolution image. The first category is to deblur or superresolve an image, then feed the restored image to the recognition engine [1, 2]. While the separated scheme is straightforward, it is not a best choice for the goal of image restoration is not consistent with that of recognition. And even worse, especially for blurred image, if the blur model is unknown or complex, notable artifacts introduced by deblurring will in fact decline the recognition performance. The second category is to do a direct recognition from blurred or low resolution image without deblurring or super resolving. Zhang et al [3] presented a joint blind restoration and recognition framework based on sparse representation. Once blur kernel is estimated, it is used to blur the training set to generate a blur dictionary and the sparse coding of the blurred face using the blur dictionary is determined to give recognition result. Moreover, the kernel is estimated iteratively in a close loop. Sun et al [4] also explored the blind blurred image recognition in which two frameworks are investigated. One is first to infer the kernel as a separate step, then the kernel is used to generate a data dictionary and an adaptive SIFT feature dictionary is also obtained accordingly. The other is to integrate the kernel estimation and the adaptive SIFT dictionary inference into a common model. The two steps are alternatively executed until stop criterion is reached. The main drawback of works in [3] and [4] is the low efficiency since the time consumption of blurring operation is heavy. Li et al [5] learned two coupled mapping matrix that mapped a pair of high and low resolution image to a unique feature space. The target of the couple mapping matrix is to make the distance between two points in feature space as close as possible provided that they are corresponding to a pair of high and low resolution version of a same image. The efficiency of the approach is high and superresolving is not necessary, but the mapped feature is global not being benefit for recognition. The last category is to extract blur invariant or insensitive features. Heikkilä et al analyzed Local Phase Quantization (LPQ) descriptor robust to centrally symmetric blur [6]. LPQ relied on short-term of Fourier transform (STFT). They noticed that the local quantized phase is nearly invariant in low frequency band. Clearly, phase information alone is not appropriate since magnitude is also useful even more important for recognition demonstrated by work [7]. Lei et al proposed Local Frequency Descriptor (LFD) that both magnitude and phase are extracted [8]. Similar to Local Binary Pattern encoding relative relations between two pixels [9], LFD is defined in terms of relations of STFT of two neighboring pixels and declared to be insensitive to arbitrary type of blur kernel.

Our idea stems from the work in [8]. It has been shown that LFD is effective for recognizing low resolution face to a certain extent. We notice that the correlations between frequencies are useful especially for improving performance of low resolution and blurred face recognition. Furthermore, for a given tested image, a descriptor that is the most insensitive relative to original image depending on the most suitable scale of STFT would be favored. To reach the two purposes, we propose Enhanced LFD and multi-scale competition strategy. The former encodes the joint information in space and frequency into the descriptor and the latter selects the winner among multi-scales relying on corresponding recognition confidences.

The paper is organized as follows: Section 2 reviews the LFD. Section 3 gives a detail discussion of the Enhanced LFD and multi-scale competition. Sections 4 demonstrates good empirical results on Yale and FERET databases. Conclusions and future work are provided in section 5.

## II. REVIEW ON LFD

LFD is based on STFT, which is calculated over a local area $N_\mathbf{x}$ centered at $\mathbf{x}$ of an image $f(\mathbf{x})$ as follows:

$$F_x(u) = \sum_{\mathbf{y}_i \in N_\mathbf{x}} f(\mathbf{y}_i)\omega^*(\mathbf{y}_i - \mathbf{x})e^{-j2\pi u^T \mathbf{y}_i} \quad (1)$$

where u = {$u_1, u_2,…,u_L$} denote a set of two dimensional frequencies, $\omega(\mathbf{x})$ denote a window function and $\omega^*(\mathbf{x})$ is the conjugate of it. A window example of size 5×5 and 4 selected frequencies are shown in Fig. 1. The STFT of an image using window of size 5×5 and the 4 selected frequencies are demonstrated in Fig. 2. In a sequel, local magnitude descriptor (*lmd*) and local phase descriptor (*lpd*) are calculated from magnitude and phase of STFT respectively. *lmd* and *lpd* are both dependent on binary strings describing relative relations between value of a position and its 8-neighborings. Once a binary string is obtained, it will be encoded into an integer. Finally, all integers in an area are pooled into a histogram.

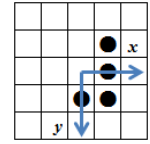

Figure 1. $u_1$=(1/5,0); $u_2$=(0,1/5); $u_3$=(1/5,1/5); $u_4$=(1/5,-1/5);

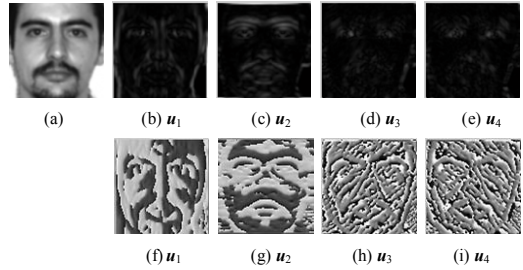

(a)   (b) $u_1$   (c) $u_2$   (d) $u_3$   (e) $u_4$

(f) $u_1$   (g) $u_2$   (h) $u_3$   (i) $u_4$

Figure 2. (a) Original face image. (b)-(e) magnitudes, (f)-(i) phases at frequencies $u_1, u_2, u_3$ and $u_4$ from left to right.

## III. ENHANCED LFD AND MULTI-SCALE COMPETITION

### 3.1 Enhanced LFD Using Joint Information in Space and Frequency

While LFD descriptor has been demonstrated to be effective for recognizing low resolution and blurred face, correlations among frequencies has not been explored since LFD only encoded the spatial neighboring relation in each single frequency plane (FP) independently. In fact, the joint representation in space and frequency is more descriptive and discriminative for recognition. Meanwhile, the property of low resolution and blur insensitive should be preserved. To accomplish the joint representation and degraded insensitive property simultaneously, we propose a new descriptor which is named Enhanced LFD that concatenates the binary relation corresponding to correlated frequencies and at the same spatial location. For the sake of a good trade-off between performance and efficiency, we choose arbitrary two frequencies from all frequencies as correlated frequencies. As mentioned in section 2, 4 frequency $u_1, u_2, u_3$ and $u_4$ are considered. Accordingly, a total of 12 2-frequency combinations are produced ($u_1,u_2$), ($u_1,u_3$), ($u_1,u_4$), ($u_2,u_1$), ($u_2,u_3$), ($u_2,u_4$), ($u_3,u_1$), ($u_3,u_2$), ($u_3,u_4$), ($u_4,u_1$), ($u_4,u_2$), ($u_4,u_3$). Of a couple of correlated frequencies, the former is principal FP, the latter is its correlated FP. For arbitrary a couple of correlated frequencies and an identical spatial location, the extended binary relations contain the 8-neighborings at the principal FP and 4-neighborings at the correlated FP as illustrated in Fig. 3, where $U_1$ denotes principal FP and $U_2$ de-notes its correlated FP. The motivation of computing features from $(u_i, u_j)$ and $(u_j, u_i)$ separately and differently is to achieve a proper trade-off between performance and the length of extended binary relation string.

Based on the magnitude of STFT $M(u, \mathbf{x})$ at $u$ and $\mathbf{x}$, the enhance lmd (*elmd*) is defined as follows:

$$T(M(u,\mathbf{k}), M(u,\mathbf{m})) = \begin{cases} 1, & \text{if } M(u,\mathbf{k}) \geq M(u,\mathbf{m}) \\ 0, & \text{otherwise} \end{cases} \quad (2)$$

where $\mathbf{k}$ denotes the focused spatial position and $\mathbf{m}$ denotes the position of one of neighbors of pixel positioned at $\mathbf{k}$. depending on the binary relations, *elmd* is encoded as an integer:

$$l_{elmd(u_p, u_c, \mathbf{k})} = \sum_{w=1}^{4} T(M(u_c, \mathbf{k}), M(u_c, \mathbf{m})) 2^{w-1} + \sum_{w=5}^{12} T(M(u_p, \mathbf{k}), M(u_p, \mathbf{m})) 2^{w-1} \quad (3)$$

where $u_p$ denotes principal FP and $u_c$ denotes its correlated FP. Similarly, based on the phase of STFT $P(u, \mathbf{x})$ at $u$ and $\mathbf{x}$, the enhance lpd (*elpd*) is defined as follows:

$$T(P(u,\mathbf{k}), P(u,\mathbf{m})) = \begin{cases} 1, & \text{if } P(u,\mathbf{k}) \text{ and } P(u,\mathbf{m}) \\ & \text{are in the same quadrant} \\ 0, & \text{otherwise} \end{cases} \quad (4)$$

depending on the binary relations, *elpd* is also encoded as an integer:

$$l_{elpd(u_p, u_c, \mathbf{k})} = \sum_{w=1}^{4} T(P(u_c, \mathbf{k}), P(u_c, \mathbf{m})) 2^{w-1} + \sum_{w=5}^{12} T(P(u_p, \mathbf{k}), P(u_p, \mathbf{m})) 2^{w-1} \quad (5)$$

The encoded integers of all positions compose a labeled image and the 12 labeled magnitude and phase images are shown in Fig. 4. Because of 12 bits length in coding, the encoded integer is a value between 0 and 4095 leading to a histogram with 4096 bins. In our experiment, each of the 24 labeled images is divided empirically into 4×4=16 non-overlapping sub-regions and a total of 16×12=192 regional label histograms are generated and concatenated into a long feature vector. Considering the both factors, the dimension of this feature vector will be 192×4096 for both magnitude and phase! Naturally, the major drawback of Enhance LFD is the substantially increased dimension compared with LFD. The extremely high dimension will introduce curse of dimensionality and make the feature unstable. We tackle the issue with a learning scheme: depending on training set, a global label histogram is obtained first. In a sequel, percentages of all bins are ordered and two bins of least percentage are combined into a new larger bin and the two percentages are summed as the percentage of the combined bin. Ordering and combinations of bins are executed alternatively and iteratively until a satisfied number of bins are achieved. The final kept and combined bins are called valid bins. During recognition, the original bins of a tested sub-region histogram will be adjusted and combined into a much lower number of bins in terms of the learned valid bins. The specific number of valid bins will be explained in experiment section.

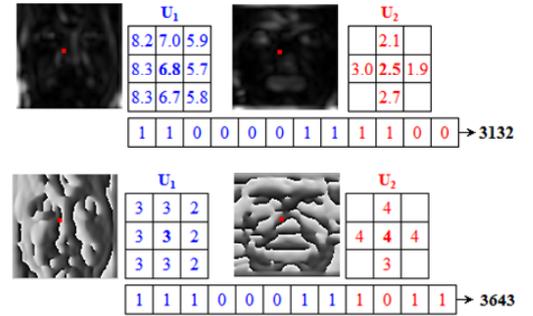

$1 \times 2^{11} + 1 \times 2^{10} + 1 \times 2^5 + 1 \times 2^4 + 1 \times 2^3 + 1 \times 2^2 = 3132$

$1 \times 2^{11} + 1 \times 2^{10} + 1 \times 2^9 + 1 \times 2^5 + 1 \times 2^4 + 1 \times 2^3 + 1 \times 2^1 + 1 \times 2^0 = 3643$

Figure 3. Enhanced LFD at a location and a couple of correlated FPs.

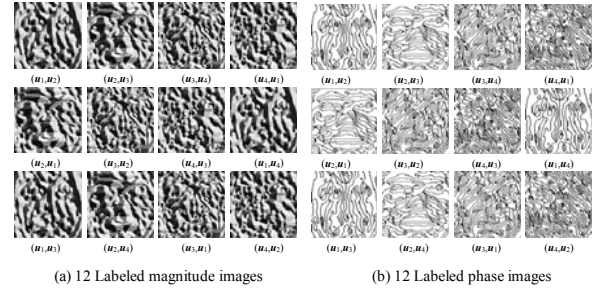

(a) 12 Labeled magnitude images     (b) 12 Labeled phase images

Figure 4. Labeled magnitude and phase images of Enhanced LFD

### 3.2 Multi-scale Competition

Another limitation of LFD is that it uses a fixed window size of STFT. In other words, it is single scale. Obviously, it is not reasonable since the degradations of the tested image will vary greatly so that the most insensitive scale corresponding to each tested image would be much different. In following, we will give an analysis about the role of scale on the recognition performance.

A low resolution or blurred image $g(\mathbf{x})$ could be modeled as a convolution between a high quality image $f(\mathbf{x})$ and a blur kernel function $k(\mathbf{x})$:

$$g(\mathbf{x}) = f(\mathbf{x}) \otimes k(\mathbf{x}) \quad (6)$$

Assume we focus on two positions $\mathbf{x}_i$ and $\mathbf{x}_j$ and two local regions centered at the two positions. In terms of STFT, the Fourier transforms of two local regions in $f(\mathbf{x})$ are as follows:

$$\begin{aligned}F_{\mathbf{x}_i}(\mathbf{u}) &= F[\omega(\mathbf{x}-\mathbf{x}_i)f(\mathbf{x})]\\ F_{\mathbf{x}_j}(\mathbf{u}) &= F[\omega(\mathbf{x}-\mathbf{x}_j)f(\mathbf{x})]\end{aligned} \quad (7)$$

where $\omega(\mathbf{x})$ refers to the window function. Now let the two local images blurred by $k(\mathbf{x})$, the deduction according to Convolution Theorem of Fourier transform is follows:

$$\begin{aligned}\tilde{G}_{\mathbf{x}_i}(\mathbf{u}) &= F\left[k(\mathbf{x}) \otimes [\omega(\mathbf{x}-\mathbf{x}_i)f(\mathbf{x})]\right] = K(\mathbf{u}) \bullet F_{\mathbf{x}_i}(\mathbf{u})\\ \tilde{G}_{\mathbf{x}_j}(\mathbf{u}) &= F\left[k(\mathbf{x}) \otimes [\omega(\mathbf{x}-\mathbf{x}_j)f(\mathbf{x})]\right] = K(\mathbf{u}) \bullet F_{\mathbf{x}_j}(\mathbf{u})\end{aligned} \quad (8)$$

where $K(\mathbf{u})$ denotes Fourier transform of $k(\mathbf{x})$. For a same frequency $\mathbf{u}_k$, the blur invariant hold due to $\tilde{G}_{\mathbf{x}_i}(\boldsymbol{u}_k)/\tilde{G}_{\mathbf{x}_j}(\boldsymbol{u}_k) = F_{\mathbf{x}_i}(\boldsymbol{u}_k)/F_{\mathbf{x}_j}(\boldsymbol{u}_k)$. Nevertheless, the blur operation is followed by local area extraction in practice which means that:

$$\begin{aligned}G_{\mathbf{x}_i}(\mathbf{u}) &= F\left[\omega(\mathbf{x}-\mathbf{x}_i)g(\mathbf{x})\right] = F\left[\omega(\mathbf{x}-\mathbf{x}_i)[k(\mathbf{x}) \otimes f(\mathbf{x})]\right]\\ G_{\mathbf{x}_j}(\mathbf{u}) &= F\left[\omega(\mathbf{x}-\mathbf{x}_j)g(\mathbf{x})\right] = F\left[\omega(\mathbf{x}-\mathbf{x}_j)[k(\mathbf{x}) \otimes f(\mathbf{x})]\right]\end{aligned} \quad (9)$$

obviously, the blur insensitive will be destroyed.

However, we can make $G_{\mathbf{x}_i}(\mathbf{u})$ and $G_{\mathbf{x}_j}(\mathbf{u})$ approximate $F_{\mathbf{x}_i}(\mathbf{u})$ and $F_{\mathbf{x}_j}(\mathbf{u})$ respectively as close as possible by letting the multiple scales compete and the winner is regarded as the most insensitive scale for a given tested degraded image. We propose a straightforward but effective strategy: in a reasonable scale range, confidences of first candidate for all possible scales are calculated and then the identity corresponding to maximum confidence is regarded as the final recognition result. This procedure is intuitively a competition among multi-scales and the scale that obtains the highest confidence would win. Certainly, features and classifiers corresponding to all scales must be extracted and constructed in advance from original high quality samples. We adopt the generalized confidence presented by [10]:

$$e(c_i \mid \boldsymbol{x}) = 1 - \frac{d_{c_i}(\boldsymbol{x})}{\min_{k \neq i} d_{c_k}(\boldsymbol{x})}$$

where $\boldsymbol{x}$ denotes a tested sample, $d_{c_i}(\boldsymbol{x})$ denotes a distance of $\boldsymbol{x}$ for category $c_i$, and $e(c_i \mid \boldsymbol{x})$ just denotes the generalized confidence of $\boldsymbol{x}$ for category $c_i$.

To prove the feasibility of the scheme, the first candidate confidences of all scales for four correctly recognized samples and four wrongly recognized samples are shown in Fig. 5 respectively. Fig. 5(a) implies that the maximum confidence or two largest confidences is significantly larger than the others for correctly recognized sample whereas the discrepancy between the maximum confidence and the others is trivial for wrongly recognized samples shown in Fig. 5(b). From another aspect illustrated in Fig. 6, we can further argue the effectiveness of the competition strategy relying on confidence. An original image and the degraded image blurred by a blur kernel shown in right-bottom part of the image are shown in Fig. 6(a). And the magnitude histograms of a sub-region emphasized by red rectangle at three different scales are illustrated in the right most columns of Fig. 6, in which the top corresponds to original image and the bottom corresponds to blur one. It could be seen that the histogram similarity between original and blur sub-region for the scale obtaining maximum confidence is the highest, that for the scale obtaining the second largest confidence is lower and that for single scale is the lowest. An evident fact is the more insensitive the descriptor is, the more similar two histograms are. Thus, this example confirms the reliability of multi-scale competition based on confidence to certain degrees.

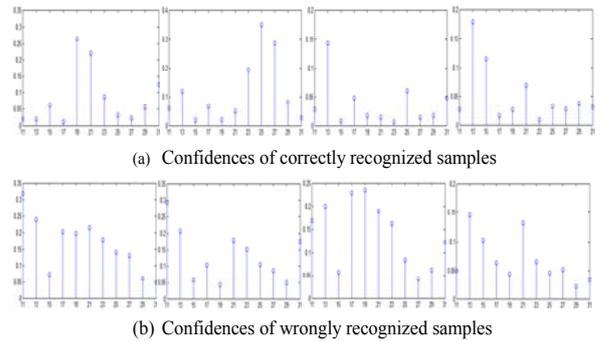

(a) Confidences of correctly recognized samples

(b) Confidences of wrongly recognized samples

Figure 5. Confidences of correctly and wrongly recognized samples

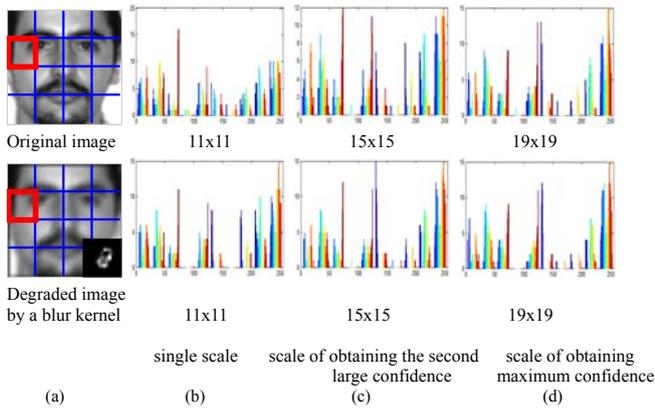

| | single scale | scale of obtaining the second large confidence | scale of obtaining maximum confidence |
|---|---|---|---|
| Original image / Degraded image by a blur kernel | 11x11 | 15x15 | 19x19 |
| (a) | (b) | (c) | (d) |

Figure 6. Magnitude histograms of a sub-region at different scales

## IV. EXPERIMENT RESULTS AND ANALYSIS

The performance of the proposed Enhanced LFD and multi-scale competition scheme are evaluated on two public face databases: Yale and FERET. FERET database used here is a random subset of original FERET containing 40 persons. All samples of each class are partitioned randomly into two parts. For Yale, one part includes five training samples and the other part includes six testing samples and for FERET, the number of training and testing sample is the same. Two low resolution degradations with down sampling scale of 2 and 4, parametric blur kernels including Gaussian kernel (standard deviation 3 and size 7×7) and linear motion kernel (7 pixel-length with 45 degrees), eight complex non-parametric kernels [11] are conducted. Altogether twelve degradations of original image in Fig. 2(a) are demonstrated in Fig. 7. Gaussian window is adopted in STFT.

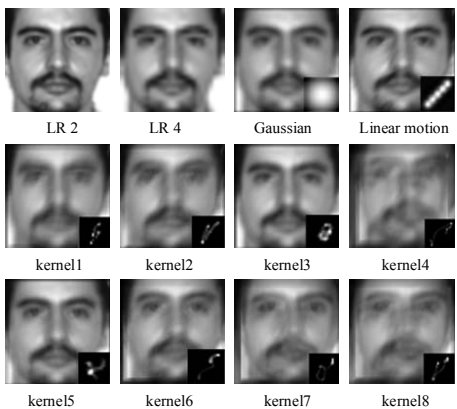

Figure 7. Twelve degradations of an original image

### 4.1 The Classifier for Face Recognition

Though we do not focus on issue of classifier in this paper, the performance of adopted classification approach is rather important. Hence, we implement a classification scheme that slightly different from [12] since we only take the reconstruction errors as recognition distance as follows:

Step 1: Calculate optimal coding $\hat{\alpha}$ for tested sample y upon dictionary D with l2-norm regularization:

$$\hat{\alpha} = \arg\min_{\alpha} \|y - D\alpha\|_2^2 + \lambda \|\alpha\|_2^2$$

where λ is the regularization factor. In all experiments, we set λ=0.01.

Step 2: Classification according to reconstruction error associated with each category:

$$\text{identity}(y) = \arg\min_i \|y - D_i \hat{\alpha}_i\|_2$$

where $\hat{\alpha}_i \in \hat{\alpha} = \{\hat{\alpha}_1, \hat{\alpha}_2, \cdots \hat{\alpha}_c\}$.

### 4.2 Parameters Setting

The single scale is set as 11×11 and the multiple scales are in range of 11×11 to 31×31. For magnitude and phase, all feature planes (for instance, 4 feature planes for LFD, 12 feature planes for Enhance LFD) are concatenated to compose a complete feature vector to feed the classifier. The optimal valid number of bins is 48 for *lmd* and *lpd* so that the dimension of LFD is $64 \times 48 = 3072$. To reduce the dimension of *elmd* and *elpd* and remain the good performance of the two enhanced descriptors as well, the optimal valid number of bins is selected as 16 for them. Consequently, the dimension of LFD and Enhance LFD are both 3072. The performance of LFD and Enhanced LFD is compared using multi-scale competition. All results have been listed in table 1 and table 2. In both tables, *lmd*/*lpd* with suffix "s" refers to single scale and "c" refers to multi-scale competition. Accordingly, *elmdc* and *elpdc* refer to *elmd* and *elpd* with multi-scale competition respectively.

TABLE I. ACCURATE RATES OF YALE (%)

|  | *lmds* | *lmdc* | *elmdc* | *lpds* | *lpdc* | *elpdc* |
|---|---|---|---|---|---|---|
| **LR2** | 98.85 | 100.00 | 100.00 | 97.70 | 97.70 | 97.70 |
| **Gaussian** | 97.70 | 100.00 | 98.85 | 66.67 | 88.51 | 88.51 |
| **motion** | 96.55 | 100.00 | 100.00 | 81.61 | 94.25 | 93.10 |
| **LR4** | 98.85 | 100.00 | 100.00 | 83.91 | 93.10 | 91.95 |
| **kernel1** | 87.36 | 97.70 | 98.85 | 35.63 | 79.31 | 79.31 |
| **kernel2** | 81.61 | 95.40 | 97.70 | 40.23 | 79.31 | 81.61 |
| **kernel3** | 96.55 | 100.00 | 98.85 | 67.82 | 89.66 | 88.51 |
| **kernel4** | 59.77 | 83.91 | 77.01 | 39.08 | 68.97 | 58.62 |
| **kernel5** | 97.70 | 100.00 | 98.85 | 74.71 | 93.10 | 91.95 |
| **kernel6** | 87.36 | 91.95 | 95.40 | 68.97 | 87.36 | 85.06 |
| **kernel7** | 85.06 | 88.51 | 95.40 | 63.22 | 81.61 | 73.56 |
| **kernel8** | 78.16 | 85.06 | 93.10 | 68.97 | 73.56 | 70.11 |
| **average** | **88.79** | **95.21** | **96.17** | **65.71** | **85.54** | **83.33** |

TABLE II. ACCURATE RATES OF FERET (%)

|  | *lmds* | *lmdc* | *elmdc* | *lpds* | *lpdc* | *elpdc* |
|---|---|---|---|---|---|---|
| **LR2** | 95.83 | 95.83 | 96.67 | 94.17 | 95.83 | 97.50 |
| **Gaussian** | 91.67 | 96.67 | 95.83 | 79.17 | 93.33 | 96.67 |
| **motion** | 94.17 | 96.67 | 96.67 | 87.50 | 96.67 | 97.50 |
| **LR4** | 90.83 | 95.00 | 95.00 | 83.33 | 95.00 | 95.83 |
| **kernel1** | 89.17 | 94.17 | 95.83 | 70.83 | 89.17 | 91.67 |
| **kernel2** | 85.00 | 95.00 | 95.00 | 72.50 | 85.00 | 91.67 |
| **kernel3** | 91.67 | 96.67 | 96.67 | 83.33 | 95.00 | 96.67 |
| **kernel4** | 51.67 | 75.83 | 75.83 | 40.00 | 70.00 | 64.17 |
| **kernel5** | 91.67 | 96.67 | 96.67 | 85.00 | 95.83 | 96.67 |
| **kernel6** | 81.67 | 92.50 | 92.50 | 75.00 | 93.33 | 91.67 |
| **kernel7** | 77.50 | 91.67 | 92.50 | 75.83 | 85.00 | 83.33 |
| **kernel8** | 70.00 | 83.33 | 90.00 | 76.67 | 81.67 | 77.50 |
| **average** | **84.24** | **92.50** | **93.26** | **76.94** | **89.65** | **90.07** |

- Single scale versus multi-scale competition

From the comparison of single scale and multi-scale competition, the great improvements achieved by the latter fully indicate that the feasibility and necessity of this strategy.

- LFD versus Enhance LFD using multi-scale competition

On average, in the scenario of multi-scale competition, results of both databases have proved the performance of *elmd* is superior to *lmd* but *elpd* is only slightly advantage over even inferior to *lpd* for FERET and Yale respectively. This may be owing to much less number of histogram bins of *elpd* and lack of discriminant analysis. This result indicates that if adequate discriminant analysis is implemented, the performance of *elpd* will surpass that of *lpd* and the advantage of *elmd* will be further increased.

V. CONCLUSIONS AND FUTURE WORK

A novel local face representation descriptor robust to low resolution and blurred degradation called Enhanced LFD and multi-scale competition strategy are proposed. Enhance LFD improves the performance of LFD by utilizing the correlations among different frequencies so as to present a joint local descriptor of two correlated frequencies at identical spatial locations. In addition, the most insensitive descriptor adaptive to tested image is found in recognition by using multi-scale competition and a depth discussion about it is presented. Encouraging results have been obtained on public obtainable Yale and FERET database.

Future work would complement discriminate analysis for the proposed descriptor instead of direct use and develop faster and cleverer searching approaches instead of full searching of multiple scales.